\documentclass{article}

\usepackage{soul}
\usepackage{amsmath}
\usepackage{microtype}
\usepackage{graphicx}
\usepackage{subfig}
\usepackage{booktabs} % for professional tables

\usepackage{hyperref}

\usepackage[accepted]{sysml2019}

\sysmltitlerunning{Exascale Deep Learning for Scientific Inverse Problems}

\begin{document}

\twocolumn[
\sysmltitle{Exascale Deep Learning for Scientific Inverse Problems}

% unless you've provided the [accepted] option to the sysml2019


\sysmlsetsymbol{equal}{*}

\begin{sysmlauthorlist}
\sysmlauthor{Nouamane Laanait}{equal,ornl-cse}
\sysmlauthor{Joshua Romero}{equal,nvidia}
\sysmlauthor{Junqi Yin}{ornl-olcf}
\sysmlauthor{M. Todd Young}{ornl-cse}
\sysmlauthor{Sean Treichler}{nvidia}
\sysmlauthor{Vitalii Starchenko}{ornl-chem}
\sysmlauthor{Albina Borisevich}{ornl-cnms}
\sysmlauthor{Alex Sergeev}{uber}
\sysmlauthor{Michael Matheson}{ornl-olcf}
\end{sysmlauthorlist}

\sysmlaffiliation{ornl-cse}{Computational Sciences and Engineering Division, Oak Ridge National Laboratory, Oak Ridge, TN, USA}
\sysmlaffiliation{ornl-cnms}{Center for Nanophase Materials Sciences, Oak Ridge National Laboratory, Oak Ridge, TN, USA}
\sysmlaffiliation{ornl-olcf}{Oak Ridge Leadership Computing Facility, Oak Ridge National Laboratory, Oak Ridge, TN, USA}
\sysmlaffiliation{ornl-chem}{Chemical Sciences Division, Oak Ridge National Laboratory, Oak Ridge, TN, USA}
\sysmlaffiliation{nvidia}{NVIDIA, Santa Clara, CA, USA}
\sysmlaffiliation{uber}{Uber Technologies, Inc., Seattle, WA, USA}

\sysmlcorrespondingauthor{Nouamane Laanait}{laanaitn@ornl.gov}
\sysmlcorrespondingauthor{Joshua Romero}{joshr@nvidia.com}

\sysmlkeywords{Distributed Deep Learning, Supercomputing, Inverse Problems, Exascale Computing}

\vskip 0.3in

\begin{abstract}
We introduce novel communication strategies in synchronous distributed Deep Learning consisting of decentralized gradient reduction orchestration and computational graph-aware grouping of gradient tensors. These new techniques produce an optimal overlap between computation and communication and result in near-linear scaling (0.93) of distributed training up to 27,600 NVIDIA V100 GPUs on the Summit Supercomputer. We demonstrate our gradient reduction techniques in the context of training a Fully Convolutional Neural Network to approximate the solution of a longstanding scientific inverse problem in materials imaging. The efficient distributed training on a dataset size of 0.5 PB, produces a model capable of an atomically-accurate reconstruction of materials, and in the process reaching a peak performance of 2.15(4) EFLOPS$_{16}$. 
\end{abstract}
]


% The \sysmlEqualContribution command is standard text for equal contribution.

\printAffiliationsAndNotice{\sysmlEqualContribution} % otherwise use the standard text.

\section{Introduction}
In light of the recent successes of ever-larger Deep Neural Networks (DNN) models and data sets \cite{dai2019transformerxl}, the need for efficient distributed machine learning strategies on massively parallel systems is more significant than ever before. Various distributed deep learning approaches have been explored throughout the years ranging from Multiple Instruction Multiple Data (MIMD) programming in model-parallelism \cite{Dean:2012MP} to the Single Program Multiple Data (SPMD) used in data-parallelism, and most recently pipelining algorithms \cite{gpipe}, parallel tensor contractions \cite{mesh-tensorflow}, and task graph-based strategies \cite{hybrid-parallel}. Despite many of these advances, data-parallelism \cite{data-parallel} remains the most widely adopted distributed deep learning strategy. Data-parallelism is both broadly applicable, and its implementation is agnostic to a system's architecture, by contrast to MIMD programming. 

As a distribution strategy, data-parallelism is communication-heavy, requiring the execution of blocking communication collectives to synchronize DNN gradients throughout a training run. A sub-optimal overlap between computation and communication operations during a single training step introduces communication overheads or inefficiencies in data-parallel distributed deep learning. On small to moderate-scale systems, with 10's - 100's of GPU/TPU accelerators, these scaling inefficiencies can be difficult to detect and systematically optimize due to system noise and load variability. Note, however, that even moderate scaling inefficiencies on the order of 5-10\% accumulate across many training steps and training runs and further increase the enormous carbon footprint of deep learning and its associated environmental impact \cite{strubell2019energy}. The scaling inefficiencies of data-parallel implementations are most readily apparent on large-scale systems such as supercomputers with 1,000's-10,000's of accelerators. Here, we show that supercomputers are ideal systems to develop and test new gradient reduction strategies to achieve near-linear scaling of data-parallelism.\footnote{The gradient reduction strategies we describe below have either been recently incorporated in the latest release of \texttt{Horovod} (https://github.com/horovod/horovod) (i.e. Bitvector Allreduce) or are currently in the pull-request review stage (i.e. Grouping).}

Extending data-parallelism to the massive scale of supercomputing systems is also motivated by the latter's traditional workload consisting of scientific numerical simulations \cite{Kent348}. In particular, infusing deep learning into scientific simulations to speed-up their execution and decrease their computational demands often requires approximating the solution of longstanding inverse problems with DNN. Here, we demonstrate the first step in this direction, made possible by our improved gradient reduction strategies. 

\section{Overview}
\subsection{System and Environment} All measurements reported here were carried out on the Summit supercomputer at the Oak Ridge Leadership Computing Facility, a US Department of Energy Office of Science User Facility. Summit is a system dedicated to open science with access applications in the form of peer-reviewed scientific user proposals.

The Summit system consists of 256 racks populated by IBM Power System AC922 compute nodes ($\sim$ 4600 nodes in total), each equipped with 2 IBM POWER9 CPUs and 6 NVIDIA V100 GPUs. It is ideally suited for Deep Learning workloads due to its node-local NVMe (burst buffer) and the Tensor Cores on V100 for faster low-precision operations. Within a Summit node, CPU-GPU and GPU-GPU communications are carried out over NVIDIA's NVLink interconnect, supporting a (peak) bi-directional bandwidth of 100 GB/s, where each 3 V100 GPUs are grouped in a ring topology with all-to-all connections to a POWER9 CPU. The CPU-CPU bridge consists of two NVLink connections, each with a (peak) bandwidth of 25 GB/s. Summit nodes are configured in a non-blocking fat-tree topology via a dual-rail Mellanox EDR 100G InfiniBand Interconnects. The IBM Alpine GPFS provides 2.5 TB/s aggregated I/O bandwidth, which is not enough to feed over 27,600 V100 GPUs each processing at over 0.5 GB/s, while NVMe offers a read bandwidth of 6 GB/s per node and provides a local I/O alternative which scales linearly with the numbers of compute nodes. 
All of the data we include here was collected (and reproduced) during normal operations of the Summit supercomputer and in the presence of other applications running on available computed nodes. As such, the performance we report is typical of the system.\\

\subsection{Distributed Deep Learning on Supercomputers}{
\label{subsec:summit-specs}

We focus on a data-parallelism approach to the distributed training of DNN. To date, the largest distributed DNN training was carried out by \cite{Kurth:2018} to learn a segmentation task on climate simulation data. These authors used a modified DNN segmentation model (DeepLabV3 \cite{deeplabv3}) which achieved a per GPU computing performance of 38.45 TFLOP$_{16}$, equivalently 31\% of the theoretical peak of the V100 GPU (the subscript 16 refers to float16 precision). 

One of the key innovations introduced in \cite{Kurth:2018} is a hierarchical Allreduce strategy consisting of intra-node collectives with NCCL (v2.3) and inter-node collectives with IBM's Spectrum-MPI. This communication strategy proved highly effective at reducing the ratio of communication time to compute time, and achieving a scaling efficiency of 90.7\% on 4560 Summit nodes with a sustained (peak) performance of 990 PFLOPS$_{16}$ (1.13 EFLOPS$_{16}$), but at the expense of skipping gradient synchronization/reduction every other training step.
The scaling efficiency used in this previous study and in other work using data-parallelism (including ours) is defined as the total number of inputs (i.e. images) processed during a training step as a function of computing resources (e.g. Summit nodes).\\
In subsequent sections, we describe new orchestration strategies of collectives during the gradients reduction stage, which prove to be more efficient than a hierarchical allreduce, allowing us to achieve $0.93$ scaling efficiency on 4600 nodes, and near perfect scaling efficiency ($>0.97$) on compute resources on the order of 1000's of GPUs or less.
}

\subsection{Distributed Training with Horovod}
\label{section:distributed}
\paragraph{Horovod}The optimized implementation of DNN mathematical operations in \texttt{cuDNN} and their fast execution on state of the art GPUs such as the V100 Tensor Cores leads to small computation times, $t_{\textrm{comp}}$, during a training step (typically $t_{\textrm{comp}} \sim$ sub-second to seconds). The time required to perform gradient reduction using blocking collectives, $t_{\textrm{comm}}$, therefore, is the key quantity to optimize in a data-parallel approach to distributed deep learning. 
We used \texttt{Horovod} ~\cite{sergeev2018horovod}, an open source library to perform gradient reduction across model replicas during distributed training. \texttt{Horovod} embeds allreduce operations into the \texttt{TensorFlow} computation graph of the DNN and employs efficient inter-GPU communication via the \texttt{MPI Allreduce} algorithm and/or by using the NVIDIA Collective Communications Library (\texttt{NCCL})~\cite{nccl-site}, depending on the configuration selected during installation time. Note that \texttt{Horovod} supports multiple frameworks and can be used to carry out data-parallelism on \texttt{PyTorch} \cite{pytorch} and \texttt{MXNet} \cite{mxnet}. 
\begin{figure}[h!]
  \centering
  \includegraphics[scale=0.55]{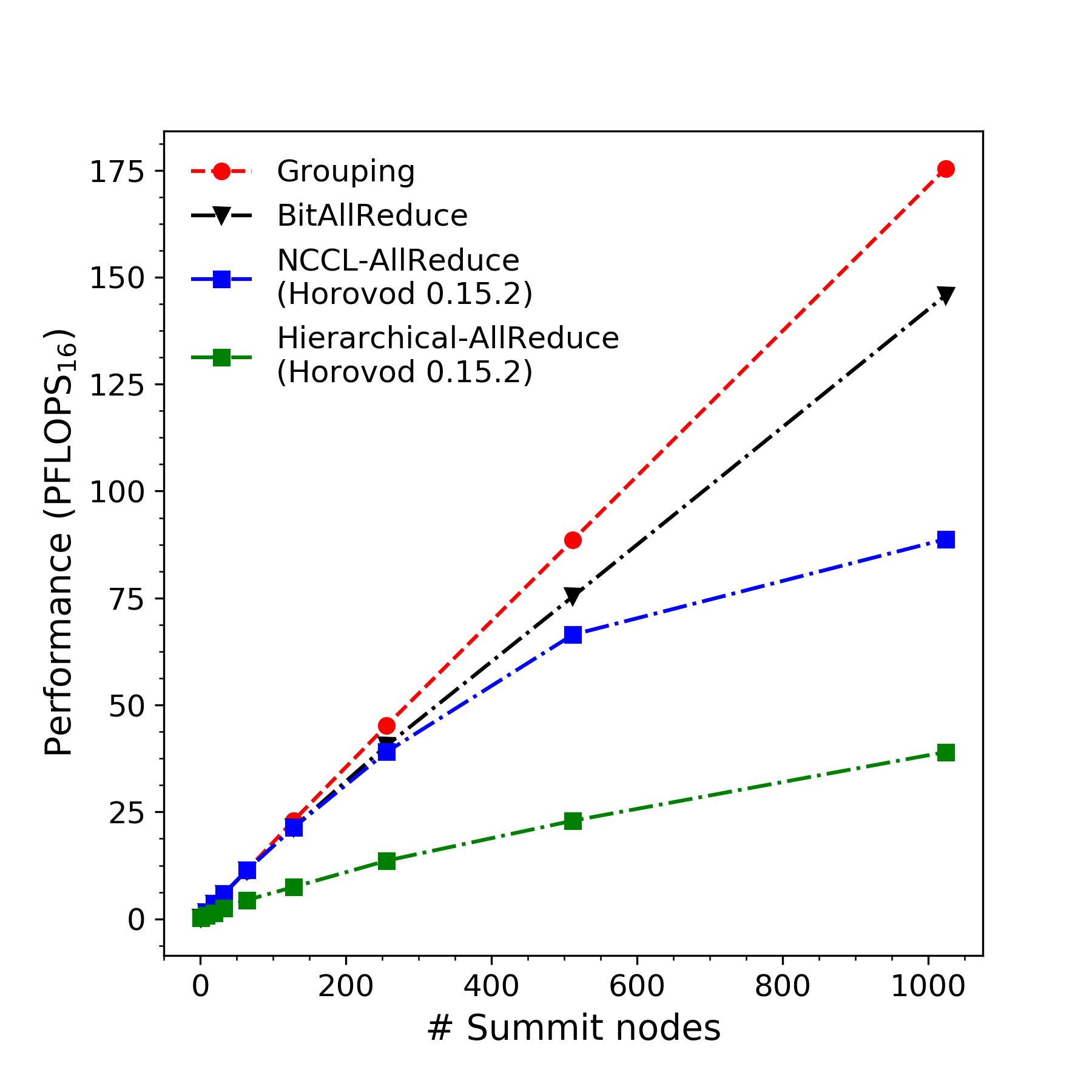}
  \caption{The influence of different Gradient Reduction Strategies on the Scaling Efficiency. The DNN model used was a modified version of the fully-convolutional dense neural network (FC-DenseNet) \cite{tiramisu} with 40 million parameters. The reported performance is the sustained performance in peta floating point operations per second carried in 16-bit numerical precision. 1 Summit node = 6 NVIDIA V100 GPUs.} \label{fig:commstrat}
\end{figure}

The hierarchical allreduce strategy introduced in \cite{Kurth:2018} was originally implemented within \texttt{Horovod} but the publicly available code base does not contain all of the features described in \cite{Kurth:2018}. As such, a direct comparison between the hierarchical allreduce in \cite{Kurth:2018} and the one we use here is not meaningful. Furthermore, some of the features of the original implementation of hierarchical allreduce made assumptions regarding the network topology that were somewhat specific to Summit's architecture.

\paragraph{Scaling of Distributed Training Strategies}In Figure \ref{fig:commstrat}, we measured the scaling efficiency of hierarchical allreduce up to 1024 Summit nodes. The sub-linear scaling is evident and was traced to poor overlap between communication and computation caused by inefficient worker coordination at large nodes. The newly released \texttt{NCCL} (v2.4) addresses latency issues of the systolic ring algorithm of \texttt{NCCL} (v2.3), using an implementation of double binary trees for full bandwidth utilization and logarithmic latency of allreduce operations \cite{Sanders:2009}. This new \texttt{NCCL} double binary trees implementation obviates the need for Horovod's explicitly hierarchical allreduce altogether, as is seen from the 3$\times$ gain in performance between the green and blue lines in \figureautorefname{\ref{fig:commstrat}}. At larger node counts ($> 1,000$ GPUs), the scaling inefficiency of data-parallelism as originally implemented in \texttt{Horovod} becomes apparent, necessitating the need for new strategies.

\section{Contributions}
\begin{figure}[h!]
  \centering
  \includegraphics[scale=0.45]{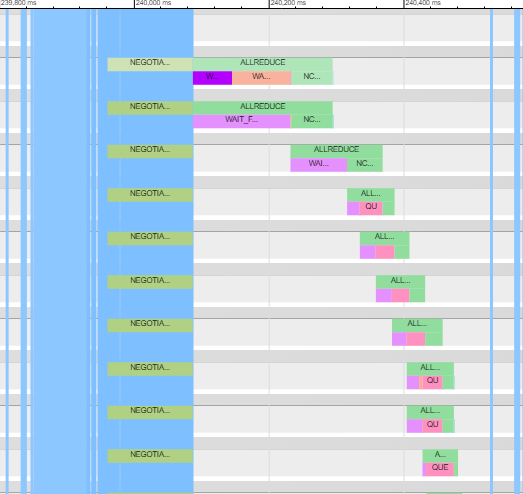}
  \caption{Horovod Timeline to illustrate the improved orchestration with grouping and bitvector allreduce. The blue vertical lines are cycle markers.}\label{fig:timeline}
\end{figure}

Our main contributions consist of:
\begin{itemize}
    \item Implementing new gradient reduction strategies which produce optimal overlap between computation and communication, a decrease in $t_{\textrm{comm}}$ during execution of the computation graph, and achieving state of the art scaling efficiency and performance of distributed deep learning up to 27,600 GPUs.
    \item Harnessing these gradient reduction strategies in the distributed training of a DNN with over $10^8$ weights on a dataset with size of a 500 TB to approximate, for the first time, a solution to an inverse problem in scientific imaging.
\end{itemize}
\paragraph{Gradient Reduction Strategies} The gradient reduction strategies consist of: (1) a lightweight worker coordination technique (BitAllReduce) and (2) a gradient tensor grouping strategy (Grouping). These two orchestration strategies improve on different aspects of distributed deep learning as currently implemented in \texttt{Horovod}. The effects of BitAllReduce and Grouping on the scaling efficiency are shown in \figureautorefname{\ref{fig:commstrat}} in black and red lines, respectively. In tandem, they lead to over $8\times$ better scaling efficiency (\figureautorefname{\ref{fig:commstrat}, \ref{fig:timeline}}). These gradient reduction strategies are computing platform agnostic and do not make any assumptions regarding the interconnect network topology.

First, Bitvector Allreduce modifies how the coordination of gradient tensors reduction via collective is performed (see Figure \ref{fig:coordination}). The main idea of Bitvector Allreduce is the use of cached meta-data, associated with each gradient tensor, and locally accessible to each \texttt{MPI}-rank to globally coordinate the execution of collective operations. In essence, we replace the original master-worker strategy of \texttt{Horovod} with a single collective (an \texttt{MPI\_Allreduce} on a bitvector) (see Figure \ref{fig:coordination:b}).

\begin{figure}
\centering
\includegraphics[scale=0.80]{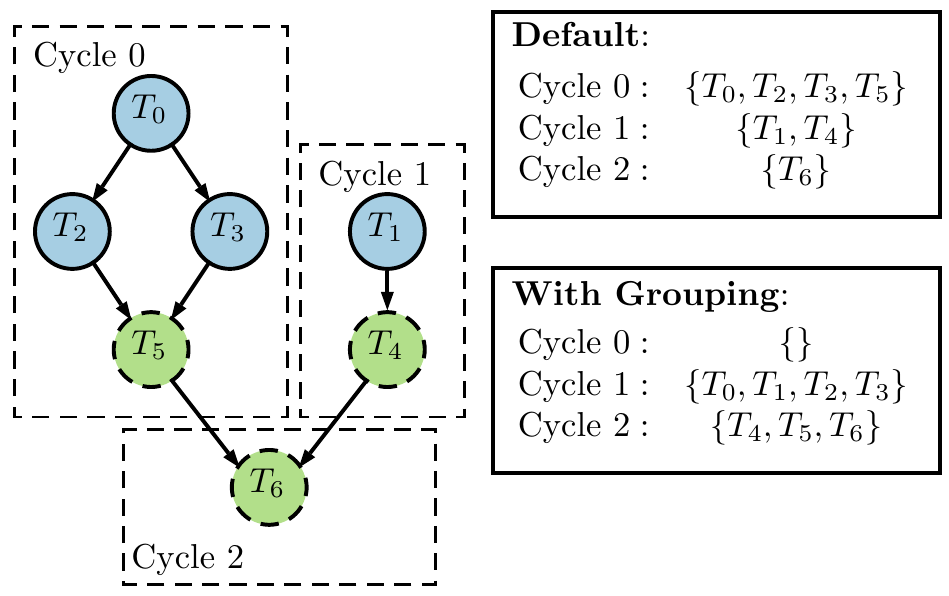}
\caption{Illustration of Grouping. A task graph with nodes that generate requests $T_n$ is depicted on the left, with the dashed boxes indicating requests visible to \texttt{Horovod} at 3 subsequent cycles. The nodes are colored to depict assignment to two groups (blue/solid borders and green/dashed borders). By default, a worker will submit all requests observed in a cycle to be processed/executed which can yield unbalanced sets of requests. With grouping enforced, requests are only submitted when complete groups are available.}
\label{fig:grouping}
\end{figure}

Second, we introduce a ``grouping" scheme for the gradient tensors akin to a graph coloring algorithm. Essentially, each \texttt{MPI} rank locally colors the nodes of its computational dependency graph (node = gradient tensor), and groups of gradient tensors are formed from like colors (see Figure \ref{fig:grouping}). Collective operations are then only issued for those groups which are ready across all ranks. One of the strengths of ``grouping" is to grant the user with the flexibility to order collectives in a fashion that exploits the architecture of her DNN model, thereby achieving greater efficiency. 

Finally, we note that both ``Grouping'' and ``Bitvector Allreduce" can be used independently, but used in combination they provided the massive gains in performance we report here. In the next section we describe in detail the implementations of these novel orchestration strategies.

\paragraph{Scientific Inverse Problems}
Harnessing the well-known function approximation capabilities of modern Deep Neural Networks (DNN) to solve challenging inverse problems in imaging \cite{Lucas:2018ev} has been mostly explored within the field of medical imaging \cite{Adler_2017, rivenson2018phase}, though there have been a few notable exceptions within materials imaging \cite{cherukara2018real, laanait2019reconstruction}. In contrast to other application domains, materials imaging, especially at the atomic scale, has the benefit of having access to highly-accurate and fully-quantitative forward simulation models and theories underpinned by quantum theory. The massive size of a single training example, which are often multi-dimensional arrays, can easily reach GBs and presents new challenges in the training of DNN. Most notably, the need for efficient I/O and the distributed training of large DNN models and consequently large message sizes. While large scale scientific simulation problems are a prevalent workload on supercomputers, to this date, however, no previous work has harnessed the capabilities of high-performance computing to produce a DNN-based solution to a scientific inverse problem. We show that our improvements to gradient reduction strategies now make it possible to approximate solutions to scientific inverse problems with deep learning and supercomputing.

\section{Gradient Reduction Strategies}
\subsection{Worker Coordination via Bitvector Allreduce}
\begin{figure}[h!]
  \centering
  \begin{subfloat}[Original coordination strategy]{
  \includegraphics[scale=0.575]{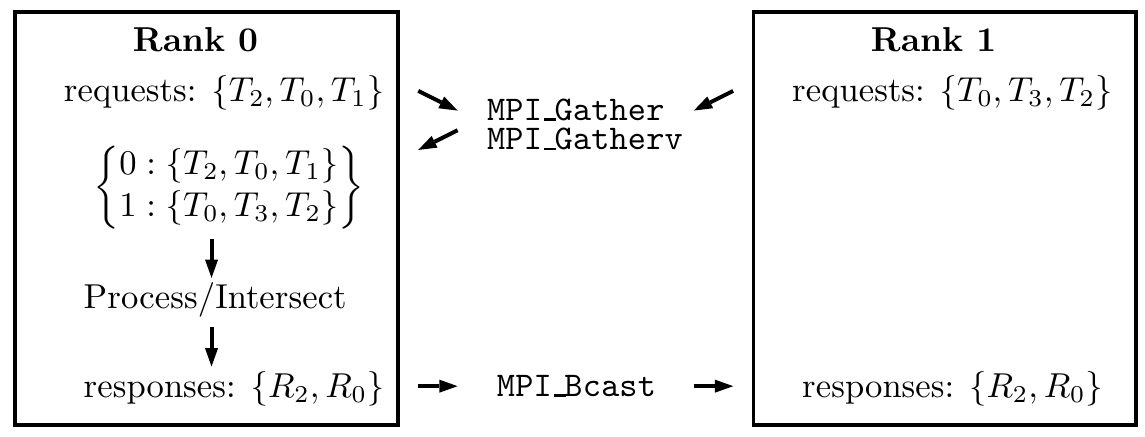}
  \label{fig:coordination:a}}
  \end{subfloat}
  \begin{subfloat}[Improved coordination strategy (bitAllReduce)]{
  \includegraphics[scale=0.525]{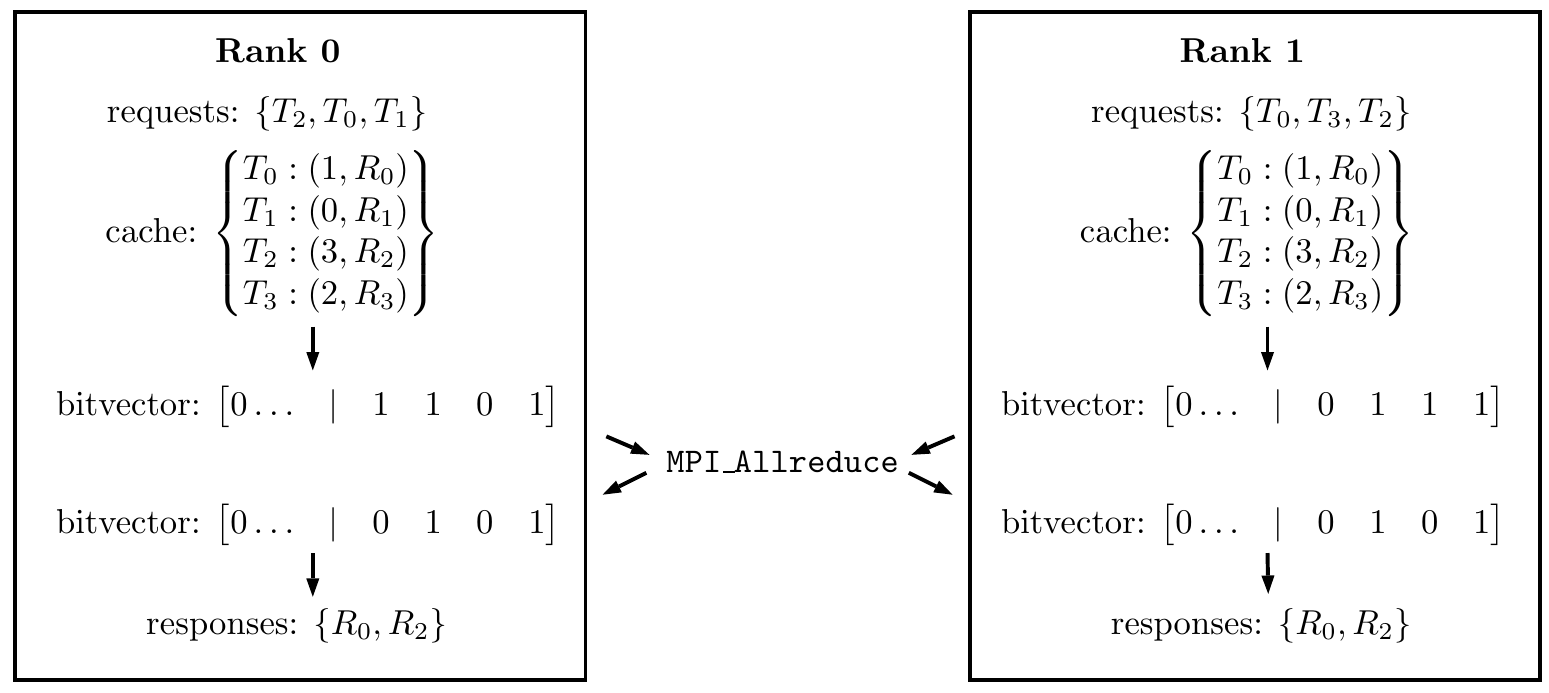}
  \label{fig:coordination:b}}
  \end{subfloat}
  \caption{Comparison of coordination strategies. \ref{fig:coordination:a}: In the original coordination strategy, Rank 0: (i) gathers requests, $T_n$, (ii) determines common requests across all ranks, (iii) forms associated responses, $R_n$, and (iv) broadcasts an ordered list of responses to all ranks for execution. \ref{fig:coordination:b}: In the improved coordination strategy, each rank checks if its requests are in the cache and sets bits in the bitvector accordingly. An initial set of bits in the bitvector are reserved for status signaling. Each cache entry is keyed by a request and maps to an integer cache bit position and stored response object. The bitvectors are globally intersected and  a list of responses associated with common set bits are obtained for execution in cache bit order.}
  \label{fig:coordination}
\end{figure}

TensorFlow's use of a graph-based scheduler permits the order of operations executed across workers to vary, even when running an identical DNN. However, collective operations which involve all workers must be performed in a globally consistent order to avoid deadlock. To solve this issue, \texttt{Horovod} introduces additional worker coordination logic to ensure all workers submit collective operations in a common order. The preexisting logic uses a master-worker coordination strategy in which a single coordinator rank is tasked with gathering \textit{requests} from all workers, determining common requests across workers, forming \textit{responses} for common requests, and then broadcasting an ordered list of responses to all workers for execution. \textit{Requests}, $T_n$, are objects submitted by each worker to request a collective operation, containing basic meta-data about the tensor involved in the operation (name, shape, datatype), as well as the type of collective operation desired (allreduce, allgather, or broadcast). \textit{Responses}, $R_n$, which are associated with a given request, contain aggregated meta-data from all workers submitting a common request (for example, all displacements for an allgather operation and the set of ranks that submitted this request), and are used for the execution of the collective operation (see Figure \ref{fig:coordination:a}). This scheme is implemented using MPI collectives, in particular \texttt{MPI\_Gatherv} and \texttt{MPI\_Bcast}, on serialized representations of the various request and response objects.

This coordination process occurs at frequent regular intervals for the duration of training, where at each tic only common collective operation requests across workers are executed. While this coordination strategy works well up to moderate scales, its effectiveness breaks down once the node count is increased further. At these larger scales, the communication cost for this coordination strategy increases to non-negligible levels, resulting in severe degradation in scaling efficiency (green and blue lines in \figureautorefname{\ref{fig:commstrat}}).

To address this, a new lightweight coordination scheme was implemented in \texttt{Horovod}, replacing the master-worker strategy and related MPI collective communication with a global intersection of a bit vector, implemented using only a single \texttt{MPI\_Allreduce} operation. One of the major overheads of the existing coordination strategy is that although identical collective operations are completed during every training iteration, requests for each operation are redundantly communicated to the coordinator rank in order to create new responses for execution.  To avoid this, we implemented a caching scheme where the responses to execute collective operations are gathered and processed by the coordinator rank only once, with the broadcasted result of this process stored in a cache on every worker. On subsequent iterations, this cached response can be directly used by each worker, bypassing redundant communication of requests to the coordinator rank. Assuming the cache remains globally consistent, it also forms the basis for a simple global enumeration of the collective operations and leads naturally to a simple procedure for worker coordination. For a given set of requests across workers, the coordination process is as follows:
\begin{enumerate}
    \item Each worker populates a bit vector, setting bits associated with its pending requests with bit positions determined from the cache.
    \item The bit vectors are globally intersected using \texttt{MPI\_Allreduce} with the binary \texttt{MPI\_BAND} operation.
    \item Each worker searches for set bits in the intersected bit vector and forms a list of associated cache entries. This list is the common set of collective operation requests for each worker to execute.
\end{enumerate}
A depiction of this improved coordination strategy can be seen in Figure \ref{fig:coordination:b}. This new coordination strategy greatly reduces communication overheads and resulted in significant improvements to scaling efficiency, shown in the black line in \figureautorefname{\ref{fig:commstrat}}.

\subsection{Grouping}
As noted in the previous section, worker coordination in \texttt{Horovod} occurs at a fixed tic rate, referred to in \texttt{Horovod} as the cycle time (see blue vertical lines in Figure \ref{fig:timeline}). This cycle time is user configurable at run-time via an environment variable. This tic rate controls how often worker coordination occurs and pending collective requests are processed and executed. One of the major features of \texttt{Horovod} is the ability to fuse individual collective operations into single operations on larger message buffers for better network performance. Notably, the scope of this fusion is limited to the requests that are encountered during a single coordination cycle. This leads to a coupling between the cycle time and collective message sizes, where in any given iteration, a shorter cycle time will lead to a more responsive execution of many collective operations with small message sizes, while a larger cycle time will lead to a slower execution of fewer collective operations with larger message sizes. This leads to a tuning dilemma: for low-latency execution of collective operations, the cycle time should be reduced as much as possible; however, for efficient network utilization, the minimum message sizes cannot be too small. Due to this, it is challenging to find an optimal cycle time that effectively balances these requirements and achieves good scaling performance.

To weaken the coupling between the cycle time and message sizes, we implemented an additional feature into \texttt{Horovod} that enables explicit assignment of collective operations into groups. When using this feature, rather than executing all collective operation requests encountered during a given cycle, only requests forming a complete group are fused and executed. If multiple complete groups are encountered, they are fused together into larger messages. By enforcing a lower bound on fusion to complete groups only, a minimum message size independent of the cycle time is enforced. This enables the use of lower cycle time for low-latency execution with a constant minimum message size, maintaining efficient network utilization. Usage of this feature in tandem with the lightweight bitvector coordination scheme described previously, yielded the red performance curve in Figure \ref{fig:commstrat}, a significant improvement in scaling behavior.  

\section{Results}
\subsection{Power Efficiency}
A strong indicator of the efficiency of an application on a supercomputer is the measured power consumption. In particular, the use of blocking collectives such as Allreduce causes all operations executed on a GPU/CPU to cease until the result from the collectives are returned. For instance, in a case where the reduction of gradients stalls due to overheads introduced by an inefficient coordination strategy, this stalling would be reflected in the GPU power consumption via a cyclic increase and decrease in the power as a function of application run-time or equivalently, in our case, the training steps. 

\begin{figure}
\centering
\includegraphics[scale=0.05]{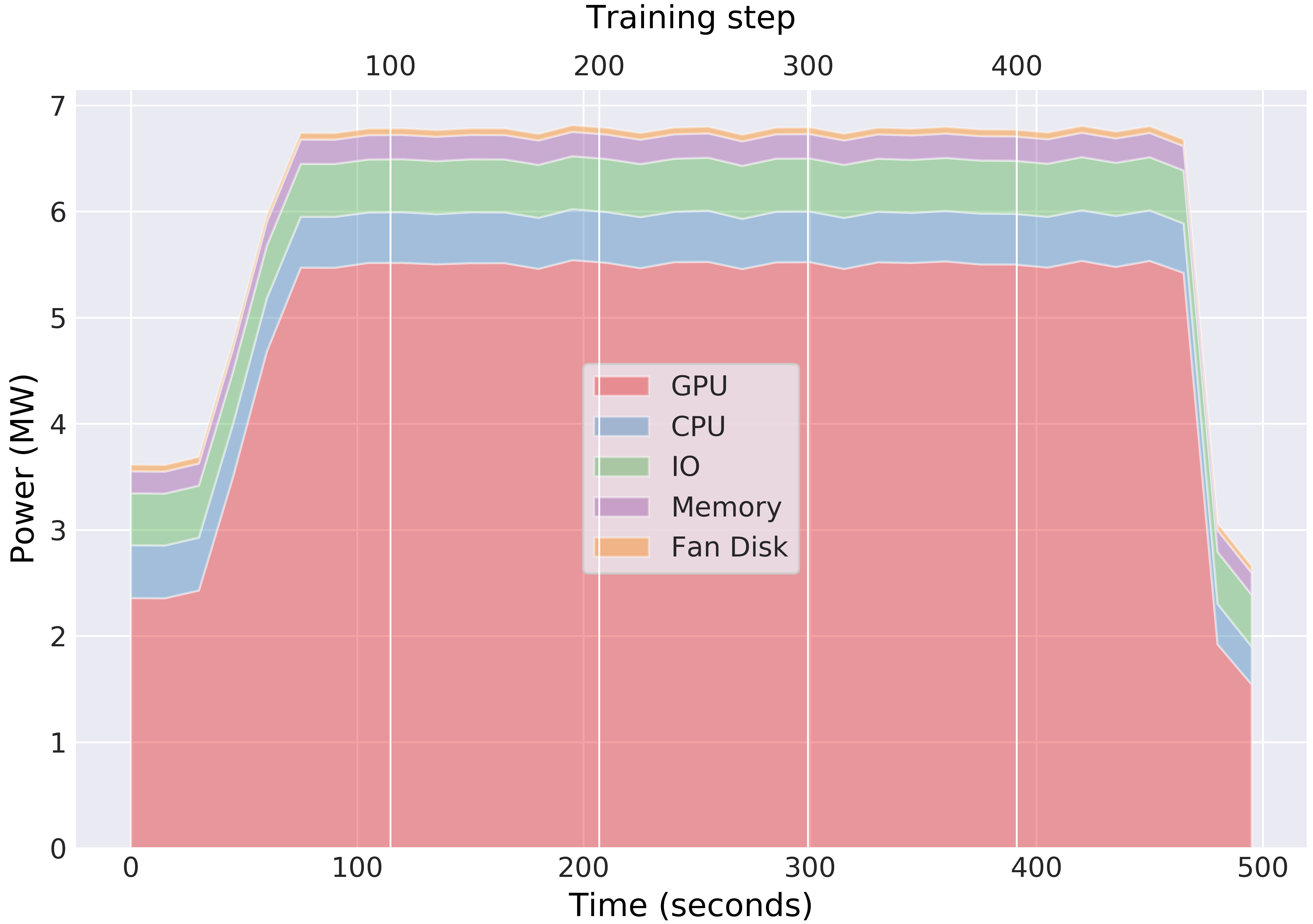}
\caption{Profiling of Summit's Power Consumption during Distributed Training on 4600 Nodes. Power profiles were collected for the main hardware components of Summit (GPU, CPU, etc...) during one of our distributed training runs. Despite the use of blocking collectives, our orchestration strategies ensure that communication and computation are optimally overlapped as reflected in a near-constant GPU power usage profile sampled at time intervals similar to the duration of a training step.}
\label{fig:power}
\end{figure}

In \figureautorefname{\ref{fig:power}}, we present the measured power consumption of the main hardware components on Summit during a distributed training run using Bitvector Allreduce and Grouping. The DNN model used in that training run and throughout the rest of the presented results is modified version of the fully-convolutional dense neural network (FC-DenseNet) \cite{tiramisu} with 220 million parameters. This choice of model produces a message size large enough to ensure that our experiments tests the robustness of the new gradient reduction strategies. The distributed training run shown in \figureautorefname{\ref{fig:power}} was carried out on 4600 out of 4608 available Summit nodes and allows us to directly measure the efficiency of our application as a whole. We found that energy metrics collected on time scales similar to the duration to a training step, show that our application's power usage is nearly constant, due to the absence of power usage fluctuations caused by GPU idleness in the presence of communication overheads.
\subsection{Performance}

\begin{figure}[h!]
  \centering
  \includegraphics[scale=0.55]{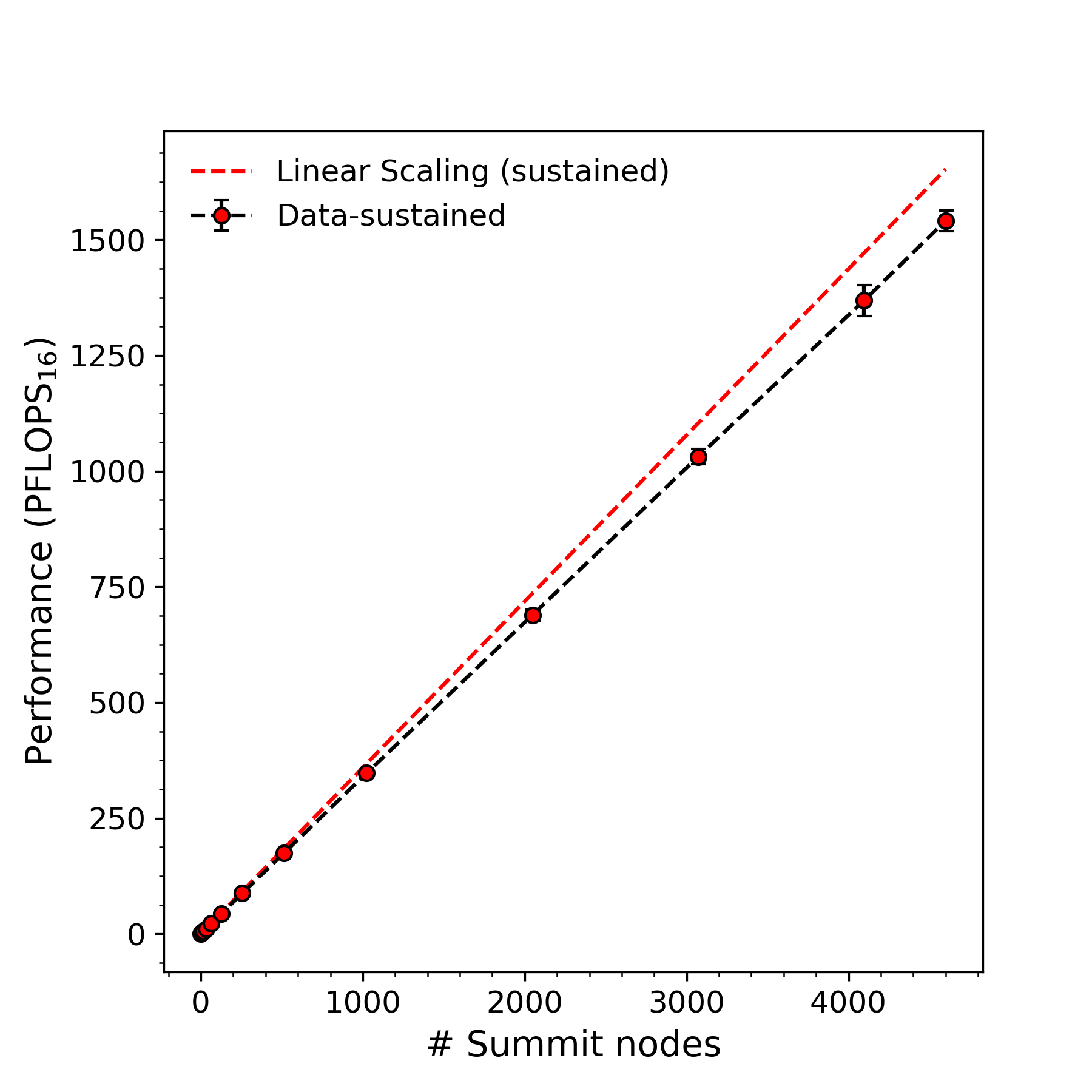}
  \caption{Scaling efficiency and Sustained Performance of distributed Deep Learning using the improved gradient reduction strategies up to 27,600 V100 GPUs.}
  \label{fig:scaling}
\end{figure}

In addition to power consumption, we also profiled the compute performance of distributed training with the new gradient reduction strategies. All of our reported performance measurements include: (1) I/O (reading of data and writing of model checkpoints), (2) computation performed for the DNN forward and backward propagation, and (3) communication operations embedded in the computation graph.

We measure the single GPU performance of our code using two distinct methods.  First, we use an analytical calculation of mathematical operations performed by DNN convolution layers assuming direct convolution.  We then augment that with the tracing of calls to \texttt{cuDNN} during execution of \texttt{TensorFlow}'s computation graph to eliminate any errors that arise from the availability of the multiple numerical implementations of the convolution operation in \texttt{cuDNN} (e.g. FFT vs. Winograd vs. direct convolution) \cite{cuDNN}. The computational complexity of these algorithms can vary substantially, and \texttt{TensorFlow} makes runtime decisions regarding which algorithm to use for each operation in the graph. As shown in \appendixautorefname{\ref{appendix}}(Table~\ref{tab:cudnn-freq}), our DNN implementation uses exclusively algorithms with a direct convolution implementation, for which the number of multiply-add operations for a direct (2-D) convolution is given by: 
\begin{equation}
 OPS_{conv} = 2 \times H \times W \times C \times K \times R \times S,
\end{equation}
where $H$,$W$ are the height and width dimensions of the inputs, $C$ and $K$ are the number of input and output channels respectively, $R$ and $S$  are the convolution kernel dimensions, and the factor of 2 accounts for ``multiply" and ``add" operations. 

The execution time of the \texttt{TensorFlow} graph,
\begin{equation}
t_{\textrm{exec}}= t_{\textrm{comm}} + t_{\textrm{comp}} + t_{\textrm{misc}},
\end{equation}
is obtained through the use of Python's built-in \texttt{time} module as well as a GPU hardware trace with \texttt{CUPTI}. The \texttt{CUPTI} trace provides the runtime of every operation individually for a single training step, whereas the application-level timing has sufficiently low overhead to be used throughout a training run. We denote the application time spent in I/O and memory copies between the host and the device as $t_{\textrm{misc}}$. $t_{\textrm{comm}}$ and $t_{\textrm{comp}}$ are the times spent on communication and computation, respectively.

The two performance numbers we report, namely sustained and peak are then given by,
\begin{equation}
\begin{aligned}
 \textrm{Sustained Performance} &= \frac{3 \times OPS_{conv}}{t_{\textrm{exec}}},\\
 \textrm{Peak Performance} &= \frac{3 \times OPS_{conv}}{t_{\textrm{comp}}},
\end{aligned}
\end{equation}
where the factor of 3 accounts for forward convolutions (Conv2D\_FWD), gradient backpropagation with respect to the convolution kernels (Conv2D\_BackpropKernel), and gradient backpropagation with respect to the inputs (Conv2D\_BackpropInput).

\subsection{Scaling}
Performance measurements on multiple nodes are carried out in a similar fashion, with the minor addition of averaging $t_{\textrm{exec}}$ across all MPI-ranks. The sustained performances reported at each node count is averaged across a distributed training run lasting 1000 steps and the variance is reported as error bars. While our definition of the peak performance at a single node does not account for $t_{\textrm{comp}}$, when we report its value on multiple nodes (see below), we multiply its value by the measured scaling efficiency ( $< 1$ for Summit nodes $> 1024$). This scaling is performed to accurately reflect the synchronous execution of our application. 

In Table \ref{tab:ops_tensorcore}, we summarize the math operations, their timing, and the overall performance during the execution of our application (one training step) on a single Summit node using the performance measurement methodology described in the previous section. We also account for the speed-up in execution enabled by the hardware implementation of half-precision intrinsics in the V100's Tensor Cores. This is done by making use of \texttt{TensorFlow}'s \texttt{TF\_DISABLE\_CUDNN\_TENSOR\_OP\_MATH} environment variable. We find that  execution with Tensor Cores produces an average speed-up of approximately $6\times$ of the computation times of the convolution operations than without (Table \ref{tab:ops_tensorcore}).

During DNN training, we attain sustained (peak) performance of 59.67 (83.92) TFLOPS$_{16}$ per GPU representing 49.7\% (70\%) of the theoretical peak of a V100 (120 TFLOPS$_{16}$), which to our knowledge, exceeds the single GPU performance of all other DNN trained on the same system to date.  

Finally, using the communication strategies described in \sectionautorefname{\ref{section:distributed}}, we are able to achieve a scaling efficiency of 0.93 at 4600 nodes during distributed deep learning (Figure \ref{fig:scaling}) and reach a sustained (peak) performance of 1.54(2) (2.15(2)) EFLOPS$_{16}$. Both our scaling efficiency and sustained performance improve significantly ($>~50\%$) on the record established by the 2018 ACM Gordon Bell prize winner \cite{Kurth:2018}. Note that in the results reported in \cite{Kurth:2018}, synchronized gradient updates were skipped every other training step, which introduces a level of asynchronous execution, and reduces their communication overhead (at the expense of gradient staleness). Our reported performance comes from fully-synchronous training, making the two results not directly comparable.

\begin{table*}
  \centering
  \caption{Math Operations, Communication, and Timing of DNN training step on 1 Summit node}
  \label{tab:ops_tensorcore}
  \setlength\tabcolsep{2 pt}
 \resizebox{\textwidth}{!}{
 \begin{tabular}{cccccc}
    \toprule
    Operation Name & Type & CUPTI Timing &  CUPTI Timing &$OPS_{conv} (\times10^{13})$  & $OPS_{conv} (\times10^{13})$ \\
    & & (ms, no Tensor Core Math) & (ms, Tensor Core Math) & (Analytical, float16) & (cuDNN , per GPU) \\
    \midrule
    Conv2D\_FWD & compute & 1244.597 & 220.801 & 1.717 & 1.717\\
    Conv2D\_BackpropKernel & compute  & 1436.822 & 226.612 & 1.717  & 1.717\\
    Conv2D\_BackpropInput & compute & 918.487 & 166.337 & 1.717  & 1.717\\ 
    NCCL-allreduce &  communication & 289.217 & 43.080 &  -  &  -\\
    (Relu, ReluGrad...) & compute & 106.911 & 107.445 & - & - \\
    MEMCPYHtoD & - & 90.542 & 99.023 & - & - \\
    \bottomrule
    & & $t_{\textrm{exec}} = 4086.576$ ms & $t_{\textrm{exec}} = 863.298$ ms & Total Math Ops = $5.151\times10^{13}$ & Sustained Performance (per GPU) = 59.67 TFLOPS$_{16}$  \\
\end{tabular}
}
\end{table*}
\section{Scientific Application}
\subsection{Problem Definition}
{
In a general inverse problem in imaging, we seek to reconstruct an image $x\in X$ from a set of measurements $y\in Y$ (typically also given by an image), where $X$ and $Y$ are (Banach) spaces. The forward operator, $F$, defined by
\begin{equation}
    \label{eq:operator}
    F : X \rightarrow Y
\end{equation}
maps the space of solutions to the space of measurements. The goal of any reconstruction method is to find $x$ by solving
\begin{equation}
    \textrm{argmin}_x ||Fx -y||_p + \lambda R(x),
\end{equation}
where $||.||_p$ denotes the $p$-norm (typically, $p=1,2$), $\lambda$ is a parameter (typically $\lambda \ll 1$), and $R(.)$ is a regularization function to incorporate \emph{a priori} knowledge about $x$ that the solution ought to obey. 

\begin{figure}[h]
  \centering
  \includegraphics[scale=0.65]{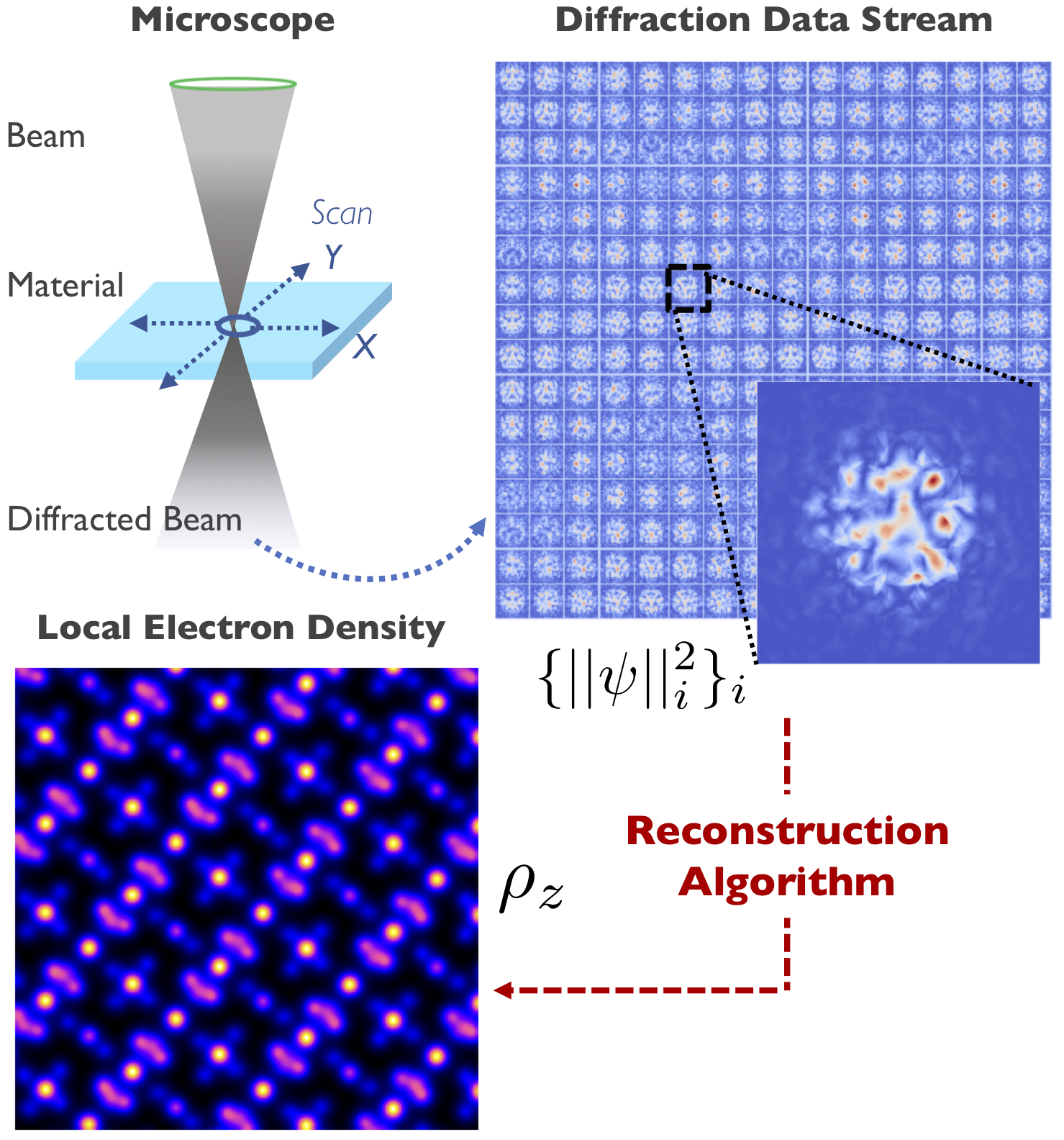}
  \caption{Reconstruction of a material's local electron density with atomic resolution from diffraction data streams acquired in an electron microscope is a longstanding inverse problem without a general solution.}\label{fig:1}
\end{figure}

In our inverse problem of interest, illustrated in \figureautorefname{\ref{fig:1}}, $x$ represents the local electron density of a material $\rho$, $y$ is a diffraction pattern $||\psi||^2$, and $F$ is the celebrated Schr\"{o}dinger equation of quantum mechanics. 
The central difficulty of the above inverse problem lies almost entirely in the fact that experimentally, one can only measure image intensities (i.e. diffraction patterns) of the exiting probe electrons $||\psi||^2$ and not the full complex-valued $\psi$ needed to find $\rho$ from $F$. Consequently, \emph{half of the information needed to directly invert the forward model is always missing}. A problem known as the phase problem \cite{born2013principles}. 

Here, we seek to learn the ``inverse" operator $F^{-1}: ||\psi||^2 \rightarrow \rho$, represented by a DNN, and trained using the technique of supervised learning with training data sampled from the forward model given by the fast-electron Schr\"{o}dinger equation \cite{kirkland2010advanced}.
}

\subsection{Simulation and the Deep Learning Model}
\label{sec:deep learning}
% \subsection{Simulation and Data}
\begin{figure*}
  \centering
  \includegraphics[scale=0.5]{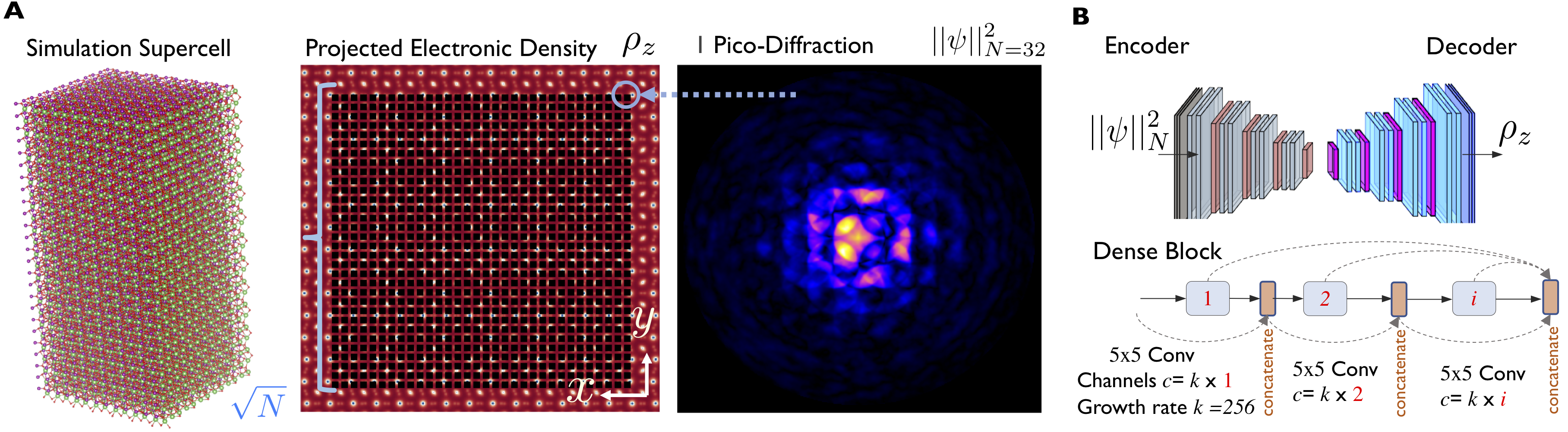}
  \caption{(A) Physics-based simulation workflow in simulating electron-wavefunction scattering through materials to generate inputs ($||\psi||^2_N$) and target outputs ($\rho_z$) of the DNN. (B) Fully-convolutional Dense Neural Net used to approximate a solution to the inverse problem of inferring $\rho_z$ from $||\psi||^2_N$.}
  \label{fig:worflow}
\end{figure*}

\paragraph{Forward Model Simulation}{
Deep Neural Networks are notoriously data hungry. To simulate enough training and test data from the forward model in optimal time, we developed a custom multi-GPU, multi-node electron scattering simulation code called \texttt{NAMSA}, which implements the multi-slice algorithm (MSA)\cite{Cowley:1957ga}, a nearly-exact solution of the fast-electron Schr\"{o}dinger equation \cite{kirkland2010advanced}.

Our simulation workflow is shown in Figure \ref{fig:worflow}A and consists of  A material supercell is built (with typical dimensions $\sim 10\times10\times20$ nm$^3$ and $\sim 100,000$ atoms), followed by a simulation of the probe electron wavefuntion interacting and propagating through all atomic planes of the supercell to produce the intensity of the exit wavefunction, $||\psi||^2$ ($512\times512$ pixels). This procedure is performed at each position on a 2-D grid (32x32) defined at the surface of the supercell. The stack of $1024 \times ||\psi||^2$ represents the inputs to our DNN,$||\psi||^2_N$, while the target outputs of the DNN is the 2-D projected electron density, $\rho_z$ ($512\times512$ pixels). The projected electron density is computed, after the scattering simulation, by integrating $\rho(\bf{r})$ along the thickness of the supercell ($z-$axis). 
}

\paragraph{Data}{
Our simulations span over 60,000 solid-state materials crystal structure files accessible via the materials project database \cite{ong2013python}. For each material structure, multiple crystallographic orientations were simulated as they produce markedly different pico-diffraction patterns and projected electron densities. In total, 400,000 configuration files were generated and then partitioned into a 90/5/5 split for training, development, and test data sets. 

Simulations of training and test data sets were generated on-the-fly and stored on the node-local burst-buffer. Given our highly-optimized simulation code \texttt{NAMSA}, we found it to be more time-effective to generate the data immediately before the start of DNN training than to stage upwards of 500 TB of data (to 4600 nodes) via the global parallel filesystem- a shared resource accessible to all users. Typically, a simulation with 0.5 hours of wall-time generates about a 200 GB data set per compute node. Note, that the number of unique samples the DNN model trains on grows linearly with the numbers of GPUs used during distributed training. The entire complement of 360,000 training configuration files are only used when distributed training reaches 4600 nodes. All data I/O (file-saving during simulation, DNN model checkpointing, and data reading during DNN training/testing) was carried out via the burst buffer and used \texttt{LMDB} (in addition to Python's built-in \texttt{multiprocessing} module during the reading stage).
}

\paragraph{Neural Network Architecture}{
Encoder-Decoder networks are prevalent in computer vision tasks such as segmentation and denoising \cite{badrinarayanan2017segnet}. This style of DNN architecture learns an encoding of multidimensional input into a compressed latent representation, followed by learning a reconstruction of a multidimensional output from an encoding along the decoder path \cite{vincent2008extracting}. Encoder-decode architectures have many variations: our work adapts a fully-convolutional dense neural networks (FC-DenseNet) \cite{tiramisu}, shown in \figureautorefname{\ref{fig:worflow}}B. The two main modifications we introduce in our model consist of: (1) Large growth rates ($k$= 256) of the number of channels of the 2-D convolution layers, and (2) replacing max pooling with average pooling. The former modification is needed to give the model enough capacity to represent our input with its 1024 channels; a smaller number of channels in the first few 2-D convolutional layers would decimate most of the information encoded in $||\psi||^2_N$. The latter modification was found in earlier work to produce substantially more accurate DNN models on atomically-resolved imaging\cite{Vasudevan:2018jd}, due the inherent sparsity of these images. The output of each dense block was passed through a rectifier non-linearity (ReLU) to compute the activation, followed by a dropout layer (with probability $p=0.5$). In total, our DNN model has $22\times10^7$ weights (free parameters).
}

\paragraph{Model Implementation}{ We trained our DNN to learn a reconstruction of the (projected) electron density, $\rho_z$ by minimizing the following loss function, 
$\mathcal{L}$ given by  
\begin{equation}
\label{eq:loss}
    \mathcal{L}(\rho_z,\bar\rho_z) = \mathcal{L}_{Huber}(\rho_z,\bar\rho_z) + \epsilon R(W_i) ,	
\end{equation}
where $\mathcal{L}_{Huber}$ is the Huber loss evaluated on the true and predicted electron densities, $\rho_z$ and $\bar{\rho}_z$, respectively. We use an $L_2$-based regularization,$R$, on the weight values $W_i$ of the model with (weight-decay) coefficient $\epsilon = 10^{-4}$. We initialized the Huber loss ``cutoff" value with $\delta=10$ and decreased it during training using an exponential decay rate policy (decay rate of 0.99 every data epoch).

Due to the large DNN model and input sizes, the 16 GB memory of a V100 can only accommodate a training batch size of 1 ($||\psi||^2_N: (1024,512,512)$, $\rho_z: (1,512,512)$), even in a float16 implementation. This batch size, however, increases linearly with the scale of distributed training, reaching 27,600 at 4600 nodes. It is well established that large batch sizes adversely affect the learning dynamics of DNN trained with stochastic gradient descent. To mitigate such effects, we used a layer-wise adaptive learning rate scaling strategy (LARS), which computes a layer-wise weight update based on the $L_2$-norm of the gradients \cite{LARS}. We used LARS in conjunction with an adaptive stochastic gradient descent optimizer (Adam optimizer, $\beta_1=0.9, \beta_2=0.999$), and a staircase learning rate decay policy. Furthermore, the warm-up policy was used to linearly increase the learning rate $\eta$, from an initial value of 0.0001, to a maximum value of $ \approx N\eta$, where $N$ is the number of GPUs (MPI-ranks) participating in the distributed training of the DNN. 

Mixed-precision training has been shown to produce similar convergence behavior and accuracy to training in pure single-precision across many DL applications \cite{Child:2019vl, LARS}, as long as judicious numerical scaling strategies are applied. Here, we performed adaptive scaling of the loss before gradient computation (and application of LARS) to avoid numerical values outside of the dynamic range of float16, using the loss scaling strategies implemented in \texttt{OpenSeq2Seq}\cite{openseq2seq}. All of our deep learning code was implemented using the \texttt{TensorFlow} (v1.13) framework \cite{tensorflow}.
}

\subsection{Model Training and Validation}

\begin{figure}
  \centering
  \includegraphics[scale=0.55]{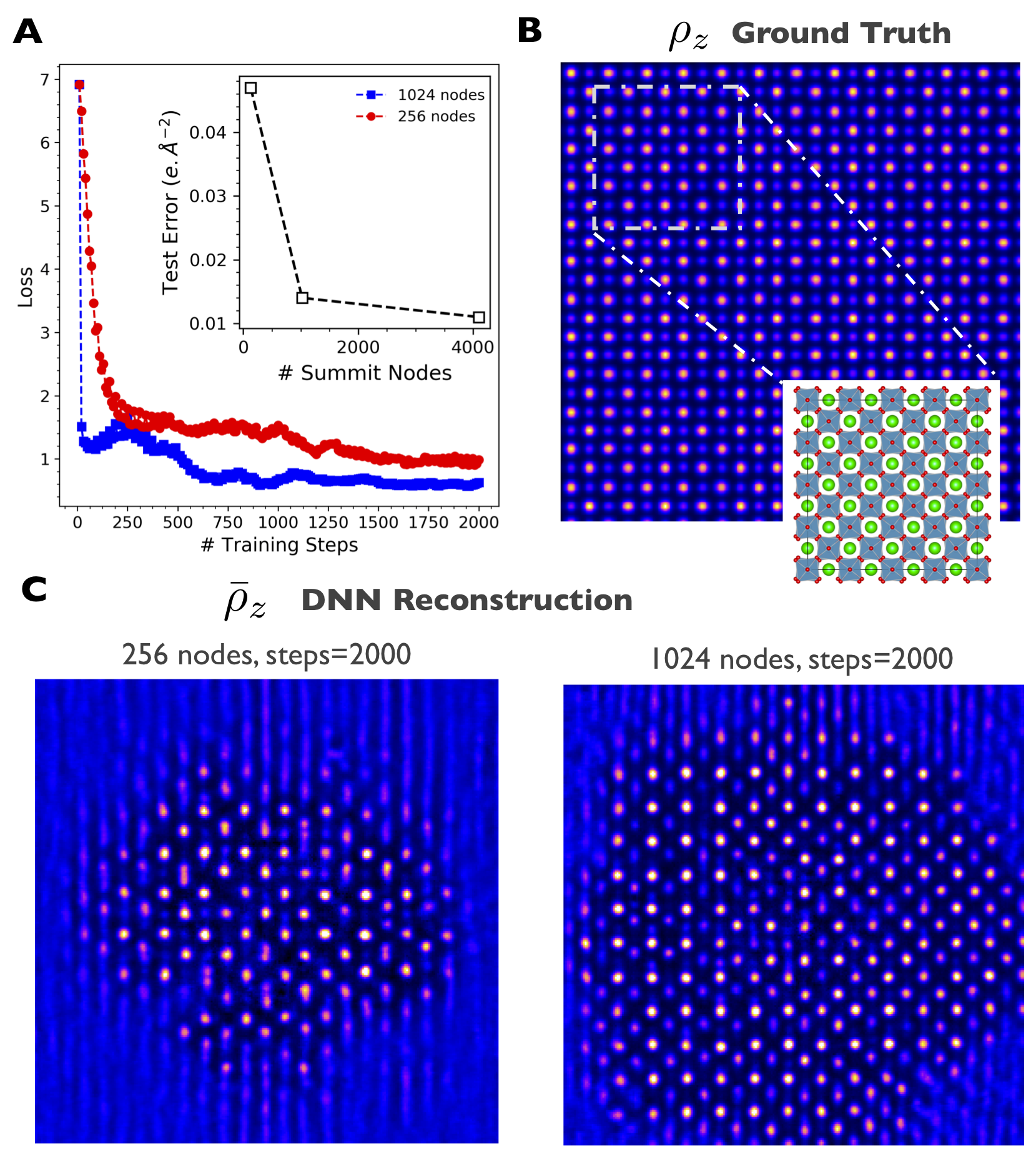}
  \caption{Comparison between the DNN-based Reconstruction and the Ground Truth Electron Density. The number of training samples processed per step is equal to the number of \texttt{MPI} ranks (6 per Summit node). The reconstruction quality was found to improve as a function of compute nodes.}
  \label{fig:predictions}
\end{figure}

We carried out multiple distributed training runs extending to 2,000 training steps. In each run, the DNN was initialized randomly and trained using the optimization strategies described in the preceding sections. We found that the training error converges reasonably well as shown in \figureautorefname{\ref{fig:predictions}} for runs spanning 128 nodes through 4096 nodes ($\approx 90\%$ of the full machine). These observations indicate that the learning strategies employed were effective in enabling good training behavior irrespective of computational scale or equivalently batch size.  

In typical data-parallelism work, the total size of the training data set, given by the number of training samples is fixed regardless of the number of DNN model replicas or equivalently the number of \texttt{MPI} ranks used in distributed training. In our application, however, the total number of unique data samples the DNN encounters during each one of the training runs depends on and grows linearly as a function of GPUs used (as discussed in \sectionautorefname{\ref{sec:deep learning}}). This linear growth in the training data set size is necessary given the finite capacity of the local node storage which can accommodate less than 1\% of the total data and the massive performance hits ($~ \times10$) our application would incur if I/O is performed directly from the larger capacity global file system (see \subsectionautorefname{\ref{subsec:summit-specs}}). 

The increase in the predictive efficacy of machine learning, deep learning in particular, as a function of growth in data is well-documented \cite{35179,sun2017revisiting}. As our data size grows as a function of \texttt{MPI}-ranks used, we expect that the quality of the DNN reconstruction on an unseen sample drawn from the test data improves. We show one such example in \figureautorefname{\ref{fig:predictions}}. We find that the reconstruction of the projected electron density is visibly far closer to the ground truth for a model trained on 1024 nodes versus 128 nodes. Both DNN models, however, fail to faithfully reconstruct the true electron density of this material across the entire field of view of the image. In the case of the DNN trained on 1024 nodes it is plausible that its reconstruction capabilities will improve with additional training and hyper-parameter tuning.

We also report the reconstruction error evaluated on the entire test data for models trained on 128, 1024, and 4096 nodes (see inset in \figureautorefname{\ref{fig:predictions}}). We find that this test error, averaged over all test samples, decreases as the number of compute (and data) increases, indicating an improving reconstruction quality on materials configurations unseen during training. 

\section{Discussion}
We have shown that by introducing new coordination strategies during gradient reductions we exceed the state of the art in scaling efficiency. This opens up, in particular, opportunities in exploiting the different levels of parallelism present in many systems (e.g. intra-node vs inter-node) such as Summit to train even larger models than we do here, for instance via the combination of model- and data-parallelism. In addition, the scaling efficiency results clearly indicate that with carefully chosen synchronized gradient reduction strategies we obtain greater utilization of the interconnect network. 

In regards to our application, the promising results shown here are a first in using DNN to solve the phase problem in the atomic imaging of materials. Future research directions can target improving the reconstruction baseline achieved here and extending the DNN-based reconstruction approach to the full 3-D electron density. Higher-dimensional reconstructions would require the use of GPU-memory intensive 3-D convolution layers presenting an additional opportunity to further benchmark the effectiveness of the novel coordination strategies we introduced here as well as extending our gradient reduction strategies to model-parallelism.

In light of the ever-increasing data streams emanating from large scale scientific user facilities, we believe this is an opportune time to harness state of the art supercomputing and machine learning. The impact of exascale machine learning on accelerating scientific progress could be, in due time, of comparable magnitude to the advances made possible via large scale physics-based simulations currently enabled by high-performance computing.

\section*{Acknowledgments}
This research was funded by a Lab Directed Research and Development project at Oak Ridge National Laboratory, a U.S. Department of Energy facility managed by UT-Battelle, LLC. An award of computer time was provided by the INCITE program. This research also used resources of the Oak Ridge Leadership Computing Facility, which is a DOE Office of Science User Facility supported under Contract DE-AC05-00OR22725. 

This manuscript has been authored by UT-Battelle, LLC under Contract No. DE-AC05-00OR22725 with the U.S. Department of Energy.
The United States Government retains and the publisher, by accepting the article for publication, acknowledges that the United States Government retains a non-exclusive, paid-up, irrevocable, world-wide license to publish or reproduce the published form of this manuscript, or allow others to do so, for United States Government purposes.
The Department of Energy will provide public access to these results of federally sponsored research in accordance with the DOE Public Access Plan (http://energy.gov/downloads/doe-public-access-plan).

%
% The next two lines define the bibliography style to be used, and the bibliography file.

\nocite{langley00}

\bibliography{biblio}
\bibliographystyle{sysml2019}

%%%%%%%%%%%%%%%%%%%%%%%%%%%%%%%%%%%%%%%%%%%%%%%%%%%%%%%%%%%%%%%%%%%%%%%%%%%%%%%
%%%%%%%%%%%%%%%%%%%%%%%%%%%%%%%%%%%%%%%%%%%%%%%%%%%%%%%%%%%%%%%%%%%%%%%%%%%%%%%
% SUPPLEMENTAL CONTENT AS APPENDIX AFTER REFERENCES
%%%%%%%%%%%%%%%%%%%%%%%%%%%%%%%%%%%%%%%%%%%%%%%%%%%%%%%%%%%%%%%%%%%%%%%%%%%%%%%
%%%%%%%%%%%%%%%%%%%%%%%%%%%%%%%%%%%%%%%%%%%%%%%%%%%%%%%%%%%%%%%%%%%%%%%%%%%%%%%
\appendix
\section{Performance}
\label{appendix}
\begin{table*}[h!]
  \centering
  \caption{Type and Frequency of cuDNN Calls during Graph Execution}
  \label{tab:cudnn-freq}
  \begin{tabular}{lccc}
    \toprule
    cuDNN Function/Algorithm & \# Calls & DNN Operation & Tensor Cores Implementation\\
    \midrule
    \texttt{CUDNN\_CONVOLUTION\_FWD\_ALGO\_IMPLICIT\_PRECOMP\_GEMM} & 480 & Forward & Yes\\
    \texttt{CUDNN\_CONVOLUTION\_FWD\_ALGO\_IMPLICIT\_GEMM} & 20 & Forward & No\\
    \texttt{CUDNN\_CONVOLUTION\_BWD\_DATA\_ALGO\_1} & 480 & Backprop & Yes\\
    \texttt{CUDNN\_CONVOLUTION\_BWD\_FILTER\_ALGO\_1} & 500 & Backprop & Yes\\ 
  \bottomrule
\end{tabular}
\end{table*}

%%%%%%%%%%%%%%%%%%%%%%%%%%%%%%%%%%%%%%%%%%%%%%%%%%%%%%%%%%%%%%%%%%%%%%%%%%%%%%%
%%%%%%%%%%%%%%%%%%%%%%%%%%%%%%%%%%%%%%%%%%%%%%%%%%%%%%%%%%%%%%%%%%%%%%%%%%%%%%%

\end{document}